\documentclass[10pt, a4paper]{article}
\usepackage{lrec}
\usepackage{multibib}
\newcites{languageresource}{Language Resources}
\usepackage{graphicx}
\usepackage{tabularx}
\usepackage{soul}
\usepackage{titlesec}
\titleformat{\section}{\normalfont\large\bf\center}{\thesection.}{1em}{}
\titleformat{\subsection}{\normalfont\SmallTitleFont\bf\raggedright}{\thesubsection.}{1em}{}
\titleformat{\subsubsection}{\normalfont\normalsize\bf\raggedright}{\thesubsubsection.}{1em}{}
\renewcommand\thesection{\arabic{section}}
\renewcommand\thesubsection{\thesection.\arabic{subsection}}
\renewcommand\thesubsubsection{\thesubsection.\arabic{subsubsection}}

\usepackage{epstopdf}
\usepackage[utf8]{inputenc}

\usepackage{hyperref}
\usepackage{xstring}

\usepackage{color}

\title{PhysNLU: A Language Resource for Evaluating Natural Language Understanding and Explanation Coherence in Physics}

\name{Jordan Meadows$^{1,2}$, Zili Zhou$^1$, Andr\'e Freitas$^{1,2}$}

\address{$^{1}$Department of Computer Science, University of Manchester\\ $^{2}$Idiap Research Institute, Switzerland \\
         jordan.meadows@postgrad.manchester.ac.uk\\
         \{zili.zhou, andre.freitas\}@manchester.ac.uk\\}

\abstract{ In order for language models to aid physics research, they must first encode representations of mathematical and natural language discourse which lead to coherent explanations, with correct ordering and relevance of statements. We present a collection of datasets developed to evaluate the performance of language models in this regard, which measure capabilities with respect to sentence ordering, position, section prediction, and discourse coherence. Analysis of the data reveals equations and sub-disciplines which are most common in physics discourse, as well as the sentence-level frequency of equations and expressions. We present baselines that demonstrate how contemporary language models are challenged by coherence related tasks in physics, even when trained on mathematical natural language objectives.\\ \newline  }

\begin{document}

\maketitleabstract

\section{Introduction}

Physics literature is a form of mathematical language which is unique beyond simply domain vocabulary. How physicists use mathematics to reason and explain, separates their field fundamentally from other disciplines, including mathematics. Many of its sub-disciplines are situated between pure mathematics and engineering, while others conjoin computer science and biology, with physical methods acting as a well-travelled bridge between the formal and natural sciences. It has not been proven, for example, that smooth solutions~\cite{pizzocchero-2021,gala-2021,miller-2021} always exist for the Navier-Stokes equations (a millenium problem) despite their widespread use in simulating and engineering fluid dynamics, while biophysics demonstrates that fundamental problems in ecology and evolution
can be characterized by computational
complexity classes~\cite{ibsen2015computational}. Physics discourse serves as a universal mechanism for generating empirically falsifiable quantitative theory in the natural sciences and engineering~\cite{smith2017derivation,coffey2012langevin}, separate to both pure mathematics and any downstream field. \newline \indent Its core traits are reflected in unique literary devices and its mathematical explanations will differ to those in the formal sciences as a result. A concrete example is the physics derivation; a core explanatory or argumentative device less rigorous and more informal than mathematical proofs~\cite{meadows2021similarity,davis2019proof,kaliszyk2015formalizing}, which generally results in predictive equations relating physical quantities, rather than generating a truth value for a given conjecture (\textit{e.g.,} twin primes). Such equations are central components of physics descriptions, with natural language forming around them and their elements, and their relation to other equations through derivations. \newline \indent Mathematics as a whole, particularly logic, is less concerned with this predictive modelling of real world systems, let alone when such systems are quantum or relativistic, or both. Suggesting that mathematicians work at a level of abstraction higher than that of physicists (\textit{i.e.,} proof frameworks compared to specific derivations), Feynman famously states that ``\textit{Mathematicians are only dealing with the structure of reasoning...}". \newline \indent Within the unique sphere of physics literature, we introduce a suite of datasets which together gauge a model's proficiency in recognising whether or not a physics-related explanation is coherent. In parallel with tasks inspired by DiscoEval~\cite{chen-etal-2019-evaluation}, we aim to ``evaluate the discourse-related knowledge captured by pretrained sentence representations" in the physics domain. From the proposed data we show that modern pretrained language models are challenged by these tasks even after fine-tuning, in particular demonstrating that a recent language model~\cite{shen2021mathbert} trained on a large corpus of mathematical text, is outperformed by even vanilla BERT-Base and all popular non-mathematical language models considered in this work. \newline\smallskip
We contribute the following:

\begin{enumerate}
    \item We introduce PhysNLU; a collection of 4 core datasets related to sentence classification, ordering, and coherence of physics explanations based on related tasks~\cite{chen-etal-2019-evaluation}. Each dataset comprises explanations extracted from Wikipedia including derivations and mathematical language. We additionally present 2 parent datasets extracted from 6.3k articles related to physics, in both raw Wikipedia data, and in a form that mimics WikiText-103~\cite{merity2016pointer}, which is a popular dataset used in related work~\cite{iter2020pretraining}. PhysNLU is avaliable online{\footnote{\url{https://github.com/jmeadows17/PhysNLU}}}.
    
    \item We provide analysis of linguistic features of physics text, including insights such as sentence and example-level distribution of mathematical content across the datasets, the frequency in which explained concepts relate to physics sub-domains, and the most frequent equations in the discourse. 
    
    \item We demonstrate how the state-of-the-art does not exhibit proficient inference capabilities with respect to tasks concerning order, coherence, relative position, and classification of sentence-level physics explanations, even when approaches have been designed explicitly for mathematical language, through baselines extracted from experiments involving a selection of pretrained language models.
\end{enumerate}

\section{Task Description}

The tasks considered in this work probe model proficiency across 4 categories, originally designed for general language, but here employed specifically for physics discourse containing mathematics. Binary Sentence Ordering tests the ability of a model to recognise order at the shortest possible scale, between two sentences. Sentence Position tests this order and position recognition at a larger scale, closer to that of full paragraphs. Discourse Coherence tests whether a model can determine whether a sequence of statements in an explanation are continuous and relevant. Sentence Section Prediction tests how well a model can link individual sentences to a specific section of an explanation. Together, in our context, they evaluate the discourse-related knowledge captured by pretrained sentence representations, and physics explanation coherence with respect to order and sentence relevance. We now describe our method for data collection for each of the 6 datasets, including the 4 directly used in the forthcoming experiments for each task as described in Figure 2, while an overview of our contributions are displayed in Figure 1.

\begin{figure}[htp!]
\centering
\includegraphics[width=0.5\textwidth]{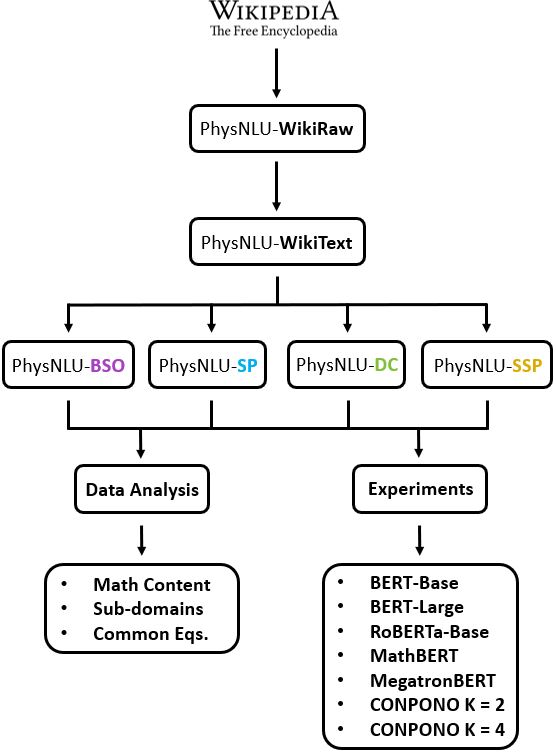}
\caption{Workflow and overview.}
\end{figure}

\begin{figure*}[htp!]
\centering
\includegraphics[width=1\textwidth]{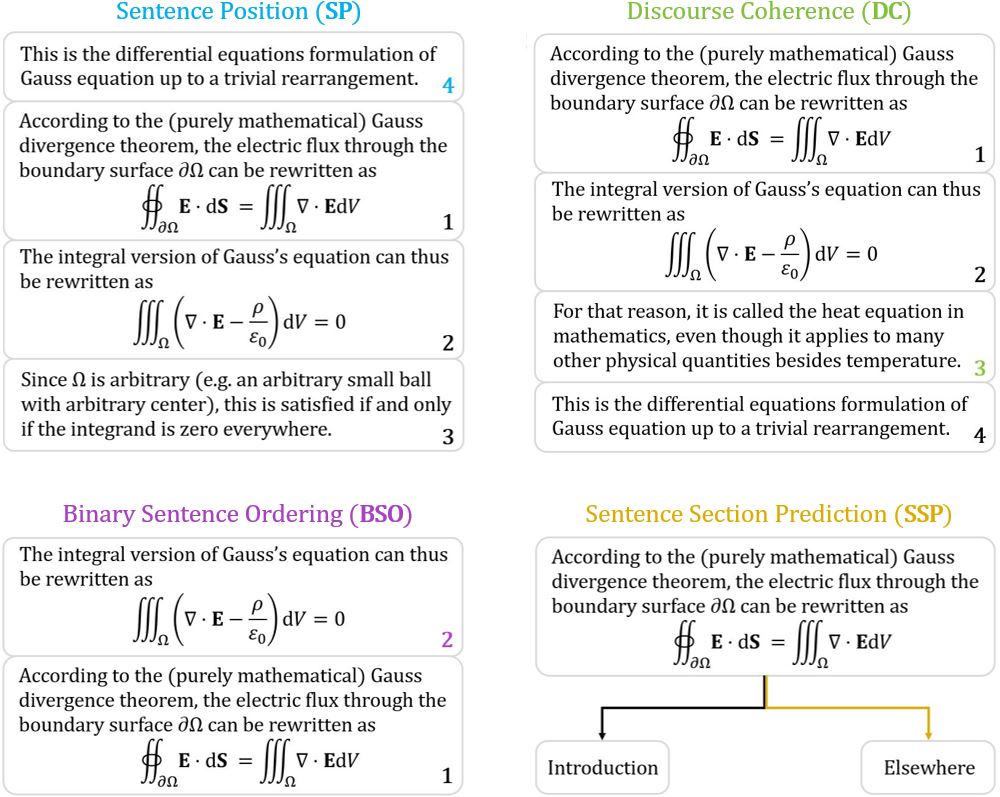}
\caption{Each numbered box is a physics statement. Explanation of how to obtain the differential form of Gauss' law via the divergence theorem, used to demonstrate how 4 evaluation tasks handle physics explanations (extracted from Wikipedia, including errors). \textbf{(SP)} Sentence position (top left) takes a random sentence from the description and moves it to position 1, then the model predicts the true position of the new first sentence, which in this case is 4. \textbf{(DC)} Discourse coherence (top right) randomly replaces a sentence with another from a different physics explanation, where the model predicts if the explanation is coherent. \textbf{(BSO)}: Binary sentence ordering (bottom left) swaps two consecutive sentences, and the model predicts if the second entails the first. \textbf{(SSP)}: Sentence section prediction takes a random sentence, and the model predicts if it belongs to an introduction or otherwise.}
\end{figure*}

\section{Dataset Collection}

\subsection{PhysNLU-WikiRaw}

Starting from an English Wikipedia XML dump, we select articles with a mention of ``physics" and contain at least one equation defined with a \texttt{<math>} tag. After cleaning articles to contain mostly mathematical natural language, and removing those which are predominantly tables, this results in a dataset containing 6.3k articles. We include article titles and corresponding raw unedited text, as well as wikipedia article categories.

\subsection{PhysNLU-WikiText}

This data mimics WikiText-103~\cite{merity2016pointer} which is used for the approach introducing the CONPONO objective~\cite{iter2020pretraining} during a preprocessing stage\footnote{\url{https://github.com/google-research/language/blob/master/language/conpono/create_pretrain_data/wiki_preproc_pipeline.py}}. Among other similarities, we opt to nest section titles within equals signs (e.g. `` = Title = ", `` = = Section = = ") and omit reference and ``see more" sections. The major linguistic differences between WikiText-103 and PhysNLU-WikiText are the inclusion of mathematical content as well as structures which may contain mathematical expressions such as tables, which are infrequent. This core dataset is taken as the starting point from which to derive the other datasets. We then extract 516k unique sentences for use in the following datasets, where sentences are determined by splits on full stops which, similarly to commas, are separated by a space from words (\textit{e.g. ``end of sentence ."}). We correct for issues with names (\textit{e.g.} J. J. Hopfield) and abbreviations, and some instances where full stops should be present but are omitted. 

\subsection{PhysNLU-BSO (Binary Sentence Ordering)}

We take all pairs of consecutive sentences from sections, where selected pairs overlap. Each pair has a 50\% chance that the pair order is swapped and we include a label to denote whether a swap (1) has occurred or not (0), suitable to be framed as binary classification. The BSO dataset contains 459k examples.

\subsection{PhysNLU-SP (Sentence Position)}

We take the first 5 sentences from each applicable section, select a sentence at random and move it to the first position (shifting the others down). The number of the swapped sentence is the label corresponding to each set of 5 sentences, suitable for multiclass classification. The SP dataset contains 40k examples. 

\subsection{PhysNLU-DC (Discourse Coherence)}

The first 6 sentences from each applicable section are selected, then between positions 2 and 5 inclusive a sentence is swapped with another article at random, with 50\% swap occurrence. Whether a swap has occurred or not is included as a label for each example for binary classification, and the DC dataset contains 35k examples.

\subsection{PhysNLU-SSP (Sentence Section Prediction)}

All sentences from the introduction sections of each article are selected and an equal number of sentences are extracted from elsewhere at random from the corpus. Introduction sentences are associated with a label (1) while non-introductory sentences are associated with a separate label (0) for binary classification. The SSP dataset contains 90k examples. 

\section{Dataset Statistics}

We now analyse our data with a focus on equations and mathematical natural language. Table 1 shows an overview of notable features, such as the proportion of examples in each dataset which contains mathematical expressions, or specifically equations. \newline \indent Figure 3 describes the proportion of sentences which contain at least $n$ mathematical elements for $n \in [1,6]$, where an element is identified via 3 separate tags: \texttt{<math>}, $\{$math$|$, and $\{$mvar$|$. The lighter bars correspond to all sentences present in the evaluation data, while the darker bars correspond to sentences in the SSP dataset which contain proportionally less math. Introductory sentences make up half of the data for SSP and usually they do not contain mathematical language or equations, which accounts for this gap. The proportion of math in sentences from the BSO, SP, and DC datasets are practically equivalent to the overall proportion. \indent Figure 4 shows the relative frequency and proportion of the top 8 Wikipedia categories associated with each article. A single article can correspond to a large number of categories, out of 12.5k categories in our case. Notably, fields related to quantum mechanics are by far the most frequent, where 10\% of the data corresponds to either ``Quantum mechanics", ``Quantum field theory", or ``Condensed matter physics".\newline \indent Figure 5 displays how often specific equations are present in the corpora. One might be tempted to claim that this demonstrates how physicists tend to argue and explain using initial conditions with respect to time ($t = 0$), displacement ($x=0$, $r=0$, $z=0$), and angle ($\theta = 0$), however this exact string matching is biased towards simple equations. As the complexity of equations increases to include multiple terms, and many terms are equivalent in meaning but different in notation, there will be multiple equations in the data which correspond to the same physics. A more accurate way to assess this would involve classifying groups of equations with a good math retrieval model~\cite{peng2021mathbert} and counting group frequency. This analysis does offer insight for simple equations however. For example, it reflects the convention that people prefer to start counting from $n=1$ in physics, which occurs more frequently than $n=0$, that the famous $E = mc^2$ is more prolific than the similarly famous $F=ma$, and that the most frequently discussed Maxwell equation is $\nabla\cdot \mathbf{B} = 0$. \newline \indent Figure 6 shows the proportion of examples from each evaluation dataset which contain at least $n$ counts of either a \texttt{<math>} equation or non-equational math, for $n \in [1, 6]$. The darker bars represent solely equations. DC, SP, BSO and SSP examples comprise 6, 5, 2, and 1 sentences respectively, so we expect that the proportion of math included decreases in that order. 

{\renewcommand{\arraystretch}{1.1}%
\begin{table}[h!]
\begin{tabular}{ c  c  c  c } 
 \hline
 \textbf{Dataset} & \textbf{Size} & \textbf{\% with math} & \textbf{\% with equations}\\ [0.1ex] 
 \hline
 DC & 35 k & 45 & 35\\
 SP & 40 k & 36 & 29\\
 BSO & 459 k & 24 & 17\\
 SSP & 90 k & 12 & 7\\
 \hline
\end{tabular}
\caption{Metadata for each evaluation dataset. The percentage of examples with math refers to the percentage of examples with LaTeX text between the XML tags \texttt{<math>} and \texttt{</math>}. The percentage of equations refers to only math which contains at least one equality.}
\label{table:1}
\end{table}}

\begin{figure}[htp!]
\centering
\includegraphics[width=0.5\textwidth]{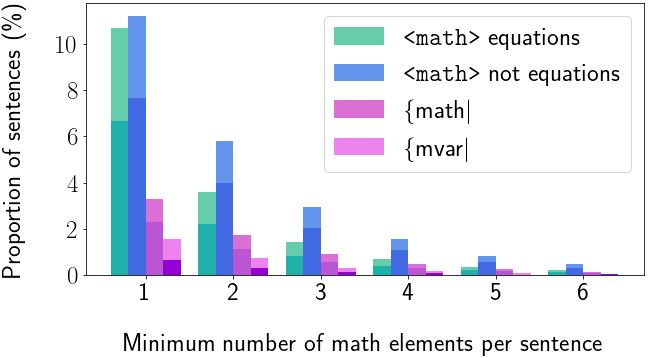}
\caption{For 516k unique sentences used in the evaluation tasks, the proportion of sentences which contain at least $n$ math elements is shown for $n \in [1,6]$. The two \texttt{<math>} variants correspond to LaTeX text identified by the XML \texttt{<math>} tag which respectively do and do not contain an equality. $\{$math$|$ and $\{$mvar$|$ correspond to any mathematical text identified by each marker. The darker bars correspond to only sentences extracted from the SSP dataset.}
\end{figure}

\begin{figure}[htp!]
\centering
\includegraphics[width=0.5\textwidth]{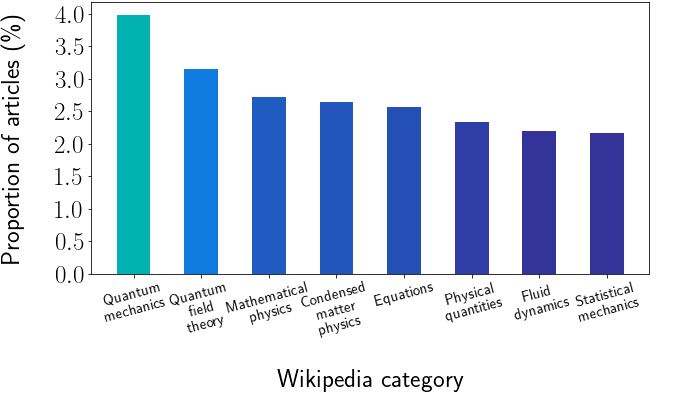}
\caption{Top 8 most frequent article categories out of 12.5k, excluding categories containing the phrase ``Articles containing ...", where each article may correspond to multiple categories.}
\end{figure}

\begin{figure}[htp!]
\centering
\includegraphics[width=0.5\textwidth]{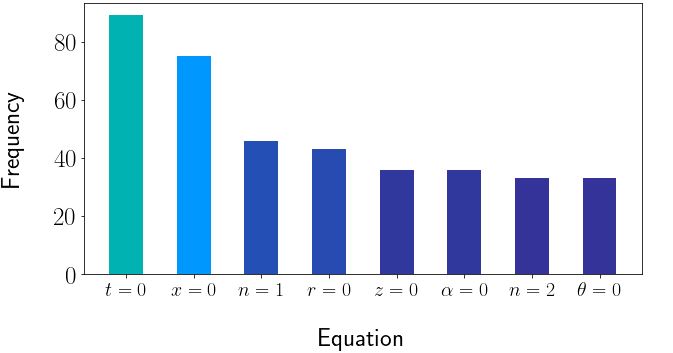}
\caption{Top 8 most frequent \texttt{<math>} tagged equations by exact string matching after accounting for spaces, commas, and full stops.}
\end{figure}

\begin{figure}[htp!]
\centering
\includegraphics[width=0.5\textwidth]{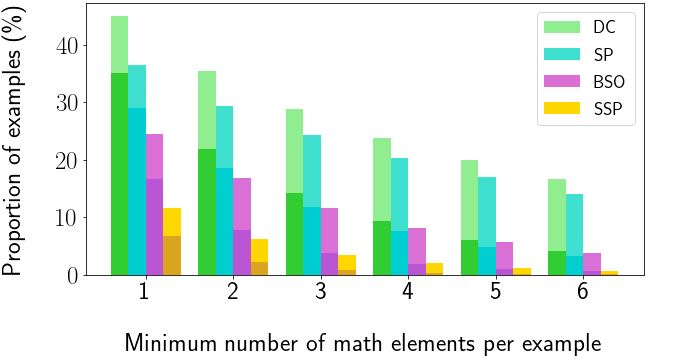}
\caption{The proportion of examples which contain at least $n$ math elements is shown for $n \in [1,6]$, where examples are sourced from the DC, SP, BSO, and SSP datasets. The math is identified via the \texttt{<math>} tag, where lighter bars correspond to at least $n$ math of any kind, while the darker bars correspond to the inclusion of only equations.}
\end{figure}

\section{Results}

We evaluate models on 4 tasks from the DiscoEval suite~\cite{chen-etal-2019-evaluation}. We remove the PDTB and RST related tasks due to the lack of a linguistic framework for describing discourse relations in the physics context. Table 1 gives additional information regarding the data used for each task. 

\subsection{Evaluation Tasks}

DiscoEval is ``designed to evaluate discourse-related knowledge in pretrained sentence representations". We briefly describe the 4 evaluation tasks from DiscoEval considered in our work, with examples shown in Figure 2. We use the same conventions~\cite{chen-etal-2019-evaluation} for representing concatenation of vectors $[.,.,...]$.\newline \indent Sentence position (SP) involves considering 5 consecutive sentences at a time, moving a random sentence to the first position, then predicting the correct position of the first sentence. We take the first 5 sentences from every paragraph in our data. Classifiers are trained by encoding the 5 sentences to vector representations $x_i$, then vectors $x_1 - x_i$ are concatenated to $x_1$ for $i \in [2,5]$ as input to the classifier as: $[x_1,\; x_1 - x_2,\; x_1 - x_3,\; x_1 - x_4,\; x_1 - x_5]$. \newline \indent Binary sentence ordering (BSO) involves taking pairs of contiguous sentences from a paragraph, swapping the order 50\% of the time and predicting if a swap has occurred. A classifier is trained by concatenating $x_1$ and $x_2$ with their element-wise difference as: $[x_1, x_2, x_1 - x_2]$. \newline \indent Discourse coherence (DC) involves taking 6 consecutive sentences, replacing a sentence from positions 2-5 inclusive with a sentence from a random article with 50\% frequency and predicting if a swap has occurred. We take the first 6 sentences from each paragraph for this task. Each vector $x_i$ is concatenated for input to the classifier as: $[x_1, x_2, x_3, x_4, x_5, x_6]$.\newline \indent Sentence section prediction (SSP) involves sampling a sentence from either the abstract of a scientific article or elsewhere with equal probability, and predicting if the sentence belongs to the abstract. In our case, we sample from article introductions as we do not have abstracts. The original task~\cite{chen-etal-2019-evaluation} involves the omission of equations which increases the difficulty of the task, but due to the nature of our problem space we leave them in. The classifier input is just the vector representation $x_1$.

\subsection{Baselines}
We include 7 baseline transformer-architecture models \cite{vaswani2017attention} in this study, BERT-base-uncased, BERT-large-uncased \cite{Devlin2019BERTPO}, RoBERTa-base \cite{Liu2019RoBERTaAR}, MathBERT \cite{shen2021mathbert}, MegatronBERT \cite{Shoeybi2019MegatronLMTM}, CONPONO (K=2), CONPONO (K=4) \cite{iter2020pretraining}.
\newline \indent BERT-base-uncased, BERT-large-uncased are two BERT models pretrained on large-scale common domain text corpora. Based on the 12 or 24 encoder layers, the models achieved state-of-the-art performances on several NLP tasks such as sentence classification, next sentence prediction, token classification, etc. BERT-large-uncased has a larger parameter size than BERT-base-uncased. The smaller model outperforms the larger in our case.
\newline \indent RoBERTa-base has the same architecture as BERT but is pretrained with more data, different hyperparameters, and only full-length sentences, and shows the BERT models were undertrained.
\newline \indent The MathBERT model uses the BERT architecture and further trains on the masked language modelling objective using a large corpus of mathematical text, including arXiv paper abstracts and textbooks covering pre-k to graduate-level. Their model shows notable improvement over BERT in an array of mathematical tasks.
\newline \indent MegatronBERT model improves the architecture of the original BERT models to enable model deployment across distributed GPU environments, while simultaneously improving the accuracy of the model.
\newline \indent The CONPONO models use the encoder architecture (and same data) from BERT to encode text segments, but are pretrained instead with the CONPONO objective together with Masked Language Modelling (MLM). The purpose is to let the models learn discourse relationships between sentences with respect to order and distance, while using negatives to increase sentence representation quality. We use 2 versions of CONPONO as baselines in this paper, K=2 means considering a maximum of 2 sentences before or after the anchor segment during pretraining, K=4 means considering a maximum of 4 sentences before or after.
\newline \indent 70-30 split testing and 5-fold split testing are conducted for each baseline. We list the results in Tables 2-5 and observe the following:
\begin{itemize}
    
    \item MathBERT is outperformed by all other models in each task, including base and large vanilla BERT. Given the improved performance of the model over BERT on 3 tasks related to mathematical text~\cite{shen2021mathbert}, and MathBERT being specialized for mathematical tasks, this is a surprising result.
    
    \item BERT-base-uncased model outperforms BERT-large-uncased in each task.
    
    \item Comparing CONPONO K=2 and CONPONO K=4, the performances are similar in each task, with K=2 being marginally better.
    
    \item MegatronBERT outperforms all models in all tasks except for discourse coherence (DC), but its parameter size is larger than most of other baselines including BERT-based-uncased and CONPONO models.
    
    \item CONPONO K=2 model outperforms BERT-base-uncased, RoBERTa-base, and MathBERT on SP and BSO tasks.
    
    \item All baselines perform poorly on the DC task.
    
    \item All baselines perform relatively well on the SSP task, but the accuracy performances are close, with no massive margin even between all-around worst performer MathBERT and generally good MegatronBERT.\newline \indent
\end{itemize}

{\renewcommand{\arraystretch}{1.3}%
\begin{table*}[htbp!]
	\centering
	\scalebox{1}{
		\begin{tabular}{@{}l|l|c|c|c|c|c|c|c c@{}}
			& \multicolumn{4}{c|}{70-30 Split} & \multicolumn{4}{c}{K-fold}\\
			\hline
			& Acc & F1 & AP & ROC & Acc & F1 & AP & ROC\\
			\hline
			BERT-base-uncased   & 0.332 & 0.320 & 0.346 & 0.676 & 0.340$\pm$0.010 & 0.314$\pm$0.007 & 0.349$\pm$0.004 & 0.678$\pm$0.004 \\
            BERT-large-uncased  & 0.306 & 0.254 & 0.318 & 0.651 & 0.306$\pm$0.006 & 0.269$\pm$0.017 & 0.317$\pm$0.004 & 0.651$\pm$0.005 \\
            RoBERTa-base        & 0.281 & 0.227 & 0.283 & 0.611 & 0.287$\pm$0.008 & 0.217$\pm$0.011 & 0.291$\pm$0.003 & 0.615$\pm$0.003 \\
            MathBERT            & 0.258 & 0.248 & 0.256 & 0.571 & 0.259$\pm$0.003 & 0.245$\pm$0.005 & 0.257$\pm$0.003 & 0.574$\pm$0.004 \\
            MegatronBERT        & 0.647 & 0.646 & 0.729 & 0.893 & 0.647$\pm$0.003 & 0.646$\pm$0.003 & 0.729$\pm$0.002 & 0.894$\pm$0.001 \\
            CONPONO K=2         & 0.536 & 0.536 & 0.577 & 0.820 & 0.532$\pm$0.005 & 0.528$\pm$0.005 & 0.574$\pm$0.006 & 0.816$\pm$0.003 \\
            CONPONO K=4         & 0.527 & 0.523 & 0.571 & 0.816 & 0.526$\pm$0.008 & 0.522$\pm$0.009 & 0.572$\pm$0.009 & 0.815$\pm$0.004 \\
		\end{tabular}
		}
	\caption{Sentence Position}
	\label{tab:SP-results}
\end{table*}}

{\renewcommand{\arraystretch}{1.3}%
\begin{table*}[htbp!]
	\centering
	\scalebox{1}{
		\begin{tabular}{@{}l|c|c|c|c|c|c|c|c@{}}
			& \multicolumn{4}{c|}{70-30 Split} & \multicolumn{4}{c}{K-fold}\\
			\hline
			& Acc & F1 & AP & ROC & Acc & F1 & AP & ROC\\
			\hline
			BERT-base-uncased   & 0.535 & 0.530 & 0.543 & 0.550 & 0.517$\pm$0.010 & 0.444$\pm$0.067 & 0.539$\pm$0.001 & 0.546$\pm$0.001 \\
            BERT-large-uncased  & 0.521 & 0.477 & 0.529 & 0.532 & 0.517$\pm$0.010 & 0.467$\pm$0.069 & 0.528$\pm$0.001 & 0.532$\pm$0.001 \\
            RoBERTa-base        & 0.531 & 0.529 & 0.537 & 0.546 & 0.531$\pm$0.002 & 0.525$\pm$0.007 & 0.538$\pm$0.001 & 0.546$\pm$0.001 \\
            MathBERT            & 0.511 & 0.443 & 0.519 & 0.518 & 0.513$\pm$0.001 & 0.492$\pm$0.023 & 0.520$\pm$0.002 & 0.520$\pm$0.002 \\
            MegatronBERT        & 0.762 & 0.762 & 0.855 & 0.853 & 0.762$\pm$0.001 & 0.762$\pm$0.001 & 0.855$\pm$0.001 & 0.853$\pm$0.001 \\
            CONPONO K=2         & 0.671 & 0.671 & 0.741 & 0.742 & 0.670$\pm$0.002 & 0.670$\pm$0.003 & 0.741$\pm$0.002 & 0.741$\pm$0.002 \\
            CONPONO K=4         & 0.665 & 0.664 & 0.734 & 0.735 & 0.666$\pm$0.001 & 0.665$\pm$0.001 & 0.734$\pm$0.001 & 0.735$\pm$0.001 \\
		\end{tabular}
	}
	\caption{Binary Sentence Ordering}
	\label{tab:BSO-results}
\end{table*}}

{\renewcommand{\arraystretch}{1.3}%
\begin{table*}[htbp!]
	\centering
	\scalebox{1}{
		\begin{tabular}{@{}l|c|c|c|c|c|c|c|c@{}}
			& \multicolumn{4}{c|}{70-30 Split} & \multicolumn{4}{c}{K-fold}\\
			\hline
			& Acc & F1 & AP & ROC & Acc & F1 & AP & ROC\\
			\hline
			BERT-base-uncased   & 0.529 & 0.527 & 0.534 & 0.541 & 0.522$\pm$0.011 & 0.491$\pm$0.044 & 0.535$\pm$0.004 & 0.539$\pm$0.004 \\
            BERT-large-uncased  & 0.505 & 0.402 & 0.519 & 0.522 & 0.514$\pm$0.006 & 0.435$\pm$0.056 & 0.520$\pm$0.007 & 0.524$\pm$0.008 \\
            RoBERTa-base        & 0.517 & 0.498 & 0.529 & 0.539 & 0.515$\pm$0.012 & 0.447$\pm$0.044 & 0.533$\pm$0.004 & 0.541$\pm$0.005 \\
            MathBERT            & 0.507 & 0.501 & 0.506 & 0.506 & 0.506$\pm$0.004 & 0.494$\pm$0.015 & 0.509$\pm$0.004 & 0.512$\pm$0.005 \\
            MegatronBERT        & 0.529 & 0.515 & 0.539 & 0.546 & 0.524$\pm$0.003 & 0.514$\pm$0.012 & 0.534$\pm$0.002 & 0.541$\pm$0.003 \\
            CONPONO K=2         & 0.534 & 0.532 & 0.538 & 0.545 & 0.533$\pm$0.003 & 0.532$\pm$0.004 & 0.536$\pm$0.003 & 0.544$\pm$0.004 \\
            CONPONO K=4         & 0.531 & 0.531 & 0.539 & 0.545 & 0.529$\pm$0.004 & 0.524$\pm$0.005 & 0.539$\pm$0.005 & 0.545$\pm$0.005 \\
		\end{tabular}
	}
	\caption{Discourse Coherence}
	\label{tab:DC-results}
\end{table*}}

{\renewcommand{\arraystretch}{1.3}%
\begin{table*}[htbp!]
	\centering
	\scalebox{1}{
		\begin{tabular}{@{}l|c|c|c|c|c|c|c|c@{}}
			& \multicolumn{4}{c|}{70-30 Split} & \multicolumn{4}{c}{K-fold}\\
			\hline
			& Acc & F1 & AP & ROC & Acc & F1 & AP & ROC\\
			\hline
			BERT-base-uncased   & 0.695 & 0.694 & 0.756 & 0.763 & 0.693$\pm$0.003 & 0.693$\pm$0.003 & 0.754$\pm$0.002 & 0.762$\pm$0.002 \\
            BERT-large-uncased  & 0.670 & 0.670 & 0.723 & 0.733 & 0.666$\pm$0.003 & 0.665$\pm$0.003 & 0.718$\pm$0.002 & 0.728$\pm$0.002 \\
            RoBERTa-base        & 0.649 & 0.648 & 0.706 & 0.713 & 0.651$\pm$0.002 & 0.650$\pm$0.002 & 0.709$\pm$0.002 & 0.715$\pm$0.002 \\
            MathBERT            & 0.618 & 0.618 & 0.679 & 0.677 & 0.619$\pm$0.006 & 0.619$\pm$0.006 & 0.677$\pm$0.003 & 0.676$\pm$0.004 \\
            MegatronBERT        & 0.705 & 0.705 & 0.778 & 0.782 & 0.704$\pm$0.003 & 0.703$\pm$0.003 & 0.776$\pm$0.001 & 0.779$\pm$0.002 \\
            CONPONO K=2         & 0.687 & 0.686 & 0.761 & 0.756 & 0.685$\pm$0.004 & 0.684$\pm$0.004 & 0.746$\pm$0.002 & 0.754$\pm$0.003 \\
            CONPONO K=4         & 0.684 & 0.683 & 0.758 & 0.755 & 0.683$\pm$0.003 & 0.683$\pm$0.003 & 0.743$\pm$0.002 & 0.752$\pm$0.002 \\
		\end{tabular}
	}
	\caption{Sentence Section Prediction}
	\label{tab:SSP-results}
\end{table*}}

\section{Related Work}

DiscoEval~\cite{chen-etal-2019-evaluation} is a suite of evaluation tasks with the purpose of determining whether sentence representations include
information about the role of a sentence in its
discourse context. They build sentence
encoders capable of modelling discourse information via training objectives that make use of natural
annotations from Wikipedia, such as nesting level, section and article titles, among others. \newline \indent Other core work~\cite{iter2020pretraining} involves pretraining on both MLM and a contrastive inter-sentence objective (CONPONO), where they achieve state-of-the-art benchmarks for five of seven tasks in the DiscoEval suite, outperforming BERT-Large despite equalling the size of BERT-Base and training on the same amount of data. BERT-Base pretrained additionally on BSO in place of CONPONO, and BERT-large, claim the remaining two benchmarks. \newline\indent Using full encoder-decoder transformer~\cite{vaswani2017attention} architecture and an additional masked attention map which incorporates relationships between nodes in operator trees (OPTs) of equations~\cite{davila2016tangent,davila2017layout}, MathBERT~\cite{peng2021mathbert} approach pretrains with three objectives on arXiv data each extracting a specific latent aspect of information. MLM learns text representations, context correspondence prediction learns the latent relationship between formula and context, and masked substructure prediction learns semantic-level structure of formulas by predicting parent and child nodes in OPTs. This model obtains state-of-the-art results in retrieval-based math tasks, but their model is unpublished. We use an alternative MathBERT~\cite{shen2021mathbert} that is trained on a larger corpus of mathematical text, and demonstrates mathematical proficiency over regular BERT.\newline \indent A data extraction pipeline~\cite{ferreira2020natural} collects 20k entries related to mathematical proofs from the ProofWiki website\footnote{\url{https://proofwiki.org/wiki/Main Page}}, such as definitions, lemmas, corollaries, and theorems. They evaluate BERT and SciBERT by fine-tuning on a pairwise relevance classification task with their NL-PS dataset, where they classify if one mathematical text is related to another. As we highlight in the Introduction, physics and mathematics literature differ in their overarching considerations, and more specifically, the unstructured informal physics Wikipedia explanations that we present in our data naturally differ from the structured proofs present in NL-PS. \newline \indent Their work builds on previous efforts applying NLP to general mathematics. One early approach~\cite{zinn2003computational} proposes proof representation structures via discourse representation theory, including a prototype for generating formal proofs from informal mathematical discourse. Another approach~\cite{cramer2009naproche} focuses on development of a controlled natural language for mathematical texts which is compatible with existing proof verification software. Since these early developments, natural language-based problem solving and theorem proving have progressed significantly, but are still below human-level performance. For example, efforts towards building datasets for evaluating math word problem solvers~\cite{huang2016well} concludes the task as a significant challenge, with more recent large-scale dataset construction and evaluation work~\cite{amini2019mathqa,miao2021diverse} confirming that model performance is still well below the gold-standard. This difficulty extends to approaches involving pre-university math problems and geometric quantities~\cite{matsuzaki2017semantic,lutheorem}. For automated theorem proving and mathematical reasoning, various datasets and accompanying approaches have been proposed~\cite{kaliszyk2017holstep,bansal2019holist} including more recent work with equational logic~\cite{piepenbrock2021learning} and language models~\cite{rabe2020mathematical,han2021proof}. \newline \indent A dataset construction approach and accompanying heuristic search for automating small physics derivations has been developed~\cite{meadows2021similarity} which allows published results in modern physics to be converted into a form interpretable by a computer algebra system~\cite{meurer2017sympy}, which then accommodates limited informal mathematical exploration. Detailed physics derivation data is scarce, and others have tackled such issues via synthetic data~\cite{aygun2020learning}, though not yet in physics. Reinforcement learning has been employed~\cite{luo2018automatic} to solve differential equations in nuclear physics with a template mapping method, and proof discovery and verification has been explored in relativity~\cite{govindarajalulu2015proof}. Theorem proving and derivation automation in physics remains elusive, with detailed discussions available in the literature~\cite{kaliszyk2015formalizing,davis2019proof}. \newline \indent With language models demonstrating logical capabilities with respect to type inference, missing assumption suggestion, and completing equalities~\cite{rabe2020mathematical}, as well as state-of-the-art performance in math retrieval and tasks related to equation-context correspondence~\cite{peng2021mathbert}, we believe our present work will contribute towards physics natural language / equational reasoners capable of generating coherent mathematical explanations and derivations.

\section{Conclusion}

Within the domain of physics, we present 2 parent datasets for general use and 4 specific datasets corresponding to discourse evaluation tasks~\cite{chen-etal-2019-evaluation} collectively referred as PhysNLU. The presented data frequently features equations, formulae, and mathematical language. Our analysis reveals that concepts related to quantum mechanics are most commonly discussed as determined by Wikipedia article category, that equations related to initial and boundary conditions are the most frequently considered when considering near-exact string matching, and we report the proportion of sentences and dataset examples which contain equations and mathematical terms identified by annotation frameworks native to Wikipedia. Finally, we present baseline results for popular non-mathematical language models and demonstrate that, despite expensive pretraining efforts and specialised training objectives for learning various aspects of mathematical text, such efforts do not improve the performance of language models in tasks related to sentence ordering, position, and recognising whether physics explanations are coherent. Future work will involve developing objectives which aid performance in this regard.

\section*{Acknowledgements}

This work was partially funded by the SNSF project NeuMath (200021\_204617).

\section{Bibliography}

\bibliographystyle{lrec}
\bibliography{lrec2020W-xample-kc}

\begin{thebibliography}{}

\bibitem[\protect\citename{Amini \bgroup et al.\egroup }2019]{amini2019mathqa}
Amini, A., Gabriel, S., Lin, P., Koncel-Kedziorski, R., Choi, Y., and
  Hajishirzi, H.
\newblock (2019).
\newblock Mathqa: Towards interpretable math word problem solving with
  operation-based formalisms.
\newblock {\em arXiv preprint arXiv:1905.13319}.

\bibitem[\protect\citename{Ayg{\"u}n \bgroup et al.\egroup
  }2020]{aygun2020learning}
Ayg{\"u}n, E., Ahmed, Z., Anand, A., Firoiu, V., Glorot, X., Orseau, L.,
  Precup, D., and Mourad, S.
\newblock (2020).
\newblock Learning to prove from synthetic theorems.
\newblock {\em arXiv preprint arXiv:2006.11259}.

\bibitem[\protect\citename{Bansal \bgroup et al.\egroup
  }2019]{bansal2019holist}
Bansal, K., Loos, S., Rabe, M., Szegedy, C., and Wilcox, S.
\newblock (2019).
\newblock Holist: An environment for machine learning of higher order logic
  theorem proving.
\newblock In {\em International Conference on Machine Learning}, pages
  454--463. PMLR.

\bibitem[\protect\citename{Chen \bgroup et al.\egroup
  }2019]{chen-etal-2019-evaluation}
Chen, M., Chu, Z., and Gimpel, K.
\newblock (2019).
\newblock Evaluation benchmarks and learning criteria for discourse-aware
  sentence representations.
\newblock In {\em Proceedings of the 2019 Conference on Empirical Methods in
  Natural Language Processing and the 9th International Joint Conference on
  Natural Language Processing (EMNLP-IJCNLP)}, pages 649--662, Hong Kong,
  China, November. Association for Computational Linguistics.

\bibitem[\protect\citename{Coffey and Kalmykov}2012]{coffey2012langevin}
Coffey, W. and Kalmykov, Y.~P.
\newblock (2012).
\newblock {\em The Langevin equation: with applications to stochastic problems
  in physics, chemistry and electrical engineering}, volume~27.
\newblock World Scientific.

\bibitem[\protect\citename{Cramer \bgroup et al.\egroup
  }2009]{cramer2009naproche}
Cramer, M., Fisseni, B., Koepke, P., K{\"u}hlwein, D., Schr{\"o}der, B., and
  Veldman, J.
\newblock (2009).
\newblock The naproche project controlled natural language proof checking of
  mathematical texts.
\newblock In {\em International Workshop on Controlled Natural Language}, pages
  170--186. Springer.

\bibitem[\protect\citename{Davila and Zanibbi}2017]{davila2017layout}
Davila, K. and Zanibbi, R.
\newblock (2017).
\newblock Layout and semantics: Combining representations for mathematical
  formula search.
\newblock In {\em Proceedings of the 40th International ACM SIGIR Conference on
  Research and Development in Information Retrieval}, pages 1165--1168.

\bibitem[\protect\citename{Davila \bgroup et al.\egroup
  }2016]{davila2016tangent}
Davila, K., Zanibbi, R., Kane, A., and Tompa, F.~W.
\newblock (2016).
\newblock Tangent-3 at the ntcir-12 mathir task.
\newblock In {\em NTCIR}.

\bibitem[\protect\citename{Davis}2019]{davis2019proof}
Davis, E.
\newblock (2019).
\newblock Proof verification technology and elementary physics.
\newblock In {\em Algorithms and Complexity in Mathematics, Epistemology, and
  Science}, pages 81--132. Springer.

\bibitem[\protect\citename{Devlin \bgroup et al.\egroup
  }2019]{Devlin2019BERTPO}
Devlin, J., Chang, M.-W., Lee, K., and Toutanova, K.
\newblock (2019).
\newblock Bert: Pre-training of deep bidirectional transformers for language
  understanding.
\newblock In {\em NAACL}.

\bibitem[\protect\citename{Ferreira and Freitas}2020]{ferreira2020natural}
Ferreira, D. and Freitas, A.
\newblock (2020).
\newblock Natural language premise selection: Finding supporting statements for
  mathematical text.

\bibitem[\protect\citename{Gala \bgroup et al.\egroup }2021]{gala-2021}
Gala, S., Galakhov, E., Ragusa, M.~A., and Salieva, O.
\newblock (2021).
\newblock Beale--kato--majda regularity criterion of smooth solutions for the
  hall-mhd equations with zero viscosity.
\newblock {\em Bulletin of the Brazilian Mathematical Society, New Series},
  pages 1--13.

\bibitem[\protect\citename{Govindarajalulu \bgroup et al.\egroup
  }2015]{govindarajalulu2015proof}
Govindarajalulu, N.~S., Bringsjord, S., and Taylor, J.
\newblock (2015).
\newblock Proof verification and proof discovery for relativity.
\newblock {\em Synthese}, 192(7):2077--2094.

\bibitem[\protect\citename{Han \bgroup et al.\egroup }2021]{han2021proof}
Han, J.~M., Rute, J., Wu, Y., Ayers, E.~W., and Polu, S.
\newblock (2021).
\newblock Proof artifact co-training for theorem proving with language models.
\newblock {\em arXiv preprint arXiv:2102.06203}.

\bibitem[\protect\citename{Huang \bgroup et al.\egroup }2016]{huang2016well}
Huang, D., Shi, S., Lin, C.-Y., Yin, J., and Ma, W.-Y.
\newblock (2016).
\newblock How well do computers solve math word problems? large-scale dataset
  construction and evaluation.
\newblock In {\em Proceedings of the 54th Annual Meeting of the Association for
  Computational Linguistics (Volume 1: Long Papers)}, pages 887--896.

\bibitem[\protect\citename{Ibsen-Jensen \bgroup et al.\egroup
  }2015]{ibsen2015computational}
Ibsen-Jensen, R., Chatterjee, K., and Nowak, M.~A.
\newblock (2015).
\newblock Computational complexity of ecological and evolutionary spatial
  dynamics.
\newblock {\em Proceedings of the National Academy of Sciences},
  112(51):15636--15641.

\bibitem[\protect\citename{Iter \bgroup et al.\egroup
  }2020]{iter2020pretraining}
Iter, D., Guu, K., Lansing, L., and Jurafsky, D.
\newblock (2020).
\newblock Pretraining with contrastive sentence objectives improves discourse
  performance of language models.

\bibitem[\protect\citename{Kaliszyk \bgroup et al.\egroup
  }2015]{kaliszyk2015formalizing}
Kaliszyk, C., Urban, J., Siddique, U., Khan-Afshar, S., Dunchev, C., and Tahar,
  S.
\newblock (2015).
\newblock Formalizing physics: automation, presentation and foundation issues.
\newblock In {\em International Conference on Intelligent Computer
  Mathematics}, pages 288--295. Springer.

\bibitem[\protect\citename{Kaliszyk \bgroup et al.\egroup
  }2017]{kaliszyk2017holstep}
Kaliszyk, C., Chollet, F., and Szegedy, C.
\newblock (2017).
\newblock Holstep: A machine learning dataset for higher-order logic theorem
  proving.
\newblock {\em arXiv preprint arXiv:1703.00426}.

\bibitem[\protect\citename{Liu \bgroup et al.\egroup }2019]{Liu2019RoBERTaAR}
Liu, Y., Ott, M., Goyal, N., Du, J., Joshi, M., Chen, D., Levy, O., Lewis, M.,
  Zettlemoyer, L., and Stoyanov, V.
\newblock (2019).
\newblock Roberta: A robustly optimized bert pretraining approach.
\newblock {\em ArXiv}, abs/1907.11692.

\bibitem[\protect\citename{Lu \bgroup et al.\egroup }2021]{lutheorem}
Lu, P., Gong, R., Jiang, S., Qiu, L., Huang, S., Liang, X., and Zhu, S.-C.
\newblock (2021).
\newblock Theorem-aware geometry problem solving with symbolic reasoning and
  theorem prediction.

\bibitem[\protect\citename{Luo and Liu}2018]{luo2018automatic}
Luo, M. and Liu, L.
\newblock (2018).
\newblock Automatic derivation of formulas using reforcement learning.
\newblock {\em arXiv preprint arXiv:1808.04946}.

\bibitem[\protect\citename{Matsuzaki \bgroup et al.\egroup
  }2017]{matsuzaki2017semantic}
Matsuzaki, T., Ito, T., Iwane, H., Anai, H., and Arai, N.~H.
\newblock (2017).
\newblock Semantic parsing of pre-university math problems.
\newblock In {\em Proceedings of the 55th Annual Meeting of the Association for
  Computational Linguistics (Volume 1: Long Papers)}, pages 2131--2141.

\bibitem[\protect\citename{Meadows and Freitas}2021]{meadows2021similarity}
Meadows, J. and Freitas, A.
\newblock (2021).
\newblock Similarity-based equational inference in physics.
\newblock {\em Physical Review Research}, 3(4), Oct.

\bibitem[\protect\citename{Merity \bgroup et al.\egroup
  }2016]{merity2016pointer}
Merity, S., Xiong, C., Bradbury, J., and Socher, R.
\newblock (2016).
\newblock Pointer sentinel mixture models.
\newblock {\em arXiv preprint arXiv:1609.07843}.

\bibitem[\protect\citename{Meurer \bgroup et al.\egroup }2017]{meurer2017sympy}
Meurer, A., Smith, C.~P., Paprocki, M., {\v{C}}ert{\'\i}k, O., Kirpichev,
  S.~B., Rocklin, M., Kumar, A., Ivanov, S., Moore, J.~K., Singh, S., et~al.
\newblock (2017).
\newblock Sympy: symbolic computing in python.
\newblock {\em PeerJ Computer Science}, 3:e103.

\bibitem[\protect\citename{Miao \bgroup et al.\egroup }2021]{miao2021diverse}
Miao, S.-Y., Liang, C.-C., and Su, K.-Y.
\newblock (2021).
\newblock A diverse corpus for evaluating and developing english math word
  problem solvers.
\newblock {\em arXiv preprint arXiv:2106.15772}.

\bibitem[\protect\citename{Miller}2021]{miller-2021}
Miller, E.
\newblock (2021).
\newblock A survey of geometric constraints on the blowup of solutions of the
  navier--stokes equation.
\newblock {\em Journal of Elliptic and Parabolic Equations}, pages 1--11.

\bibitem[\protect\citename{Peng \bgroup et al.\egroup }2021]{peng2021mathbert}
Peng, S., Yuan, K., Gao, L., and Tang, Z.
\newblock (2021).
\newblock Mathbert: A pre-trained model for mathematical formula understanding.

\bibitem[\protect\citename{Piepenbrock \bgroup et al.\egroup
  }2021]{piepenbrock2021learning}
Piepenbrock, J., Heskes, T., Janota, M., and Urban, J.
\newblock (2021).
\newblock Learning equational theorem proving.
\newblock {\em arXiv preprint arXiv:2102.05547}.

\bibitem[\protect\citename{Pizzocchero}2021]{pizzocchero-2021}
Pizzocchero, L.
\newblock (2021).
\newblock On the global stability of smooth solutions of the navier--stokes
  equations.
\newblock {\em Applied Mathematics Letters}, 115:106970.

\bibitem[\protect\citename{Rabe \bgroup et al.\egroup
  }2020]{rabe2020mathematical}
Rabe, M.~N., Lee, D., Bansal, K., and Szegedy, C.
\newblock (2020).
\newblock Mathematical reasoning via self-supervised skip-tree training.
\newblock {\em arXiv preprint arXiv:2006.04757}.

\bibitem[\protect\citename{Shen \bgroup et al.\egroup }2021]{shen2021mathbert}
Shen, J.~T., Yamashita, M., Prihar, E., Heffernan, N., Wu, X., Graff, B., and
  Lee, D.
\newblock (2021).
\newblock Mathbert: A pre-trained language model for general nlp tasks in
  mathematics education.
\newblock {\em arXiv preprint arXiv:2106.07340}.

\bibitem[\protect\citename{Shoeybi \bgroup et al.\egroup
  }2019]{Shoeybi2019MegatronLMTM}
Shoeybi, M., Patwary, M.~A., Puri, R., LeGresley, P., Casper, J., and
  Catanzaro, B.
\newblock (2019).
\newblock Megatron-lm: Training multi-billion parameter language models using
  model parallelism.
\newblock {\em ArXiv}, abs/1909.08053.

\bibitem[\protect\citename{Smith and Fleck}2017]{smith2017derivation}
Smith, R.~W. and Fleck, C.
\newblock (2017).
\newblock Derivation and use of mathematical models in systems biology.
\newblock In {\em Pollen Tip Growth}, pages 339--367. Springer.

\bibitem[\protect\citename{Vaswani \bgroup et al.\egroup
  }2017]{vaswani2017attention}
Vaswani, A., Shazeer, N., Parmar, N., Uszkoreit, J., Jones, L., Gomez, A.~N.,
  Kaiser, L., and Polosukhin, I.
\newblock (2017).
\newblock Attention is all you need.

\bibitem[\protect\citename{Zinn}2003]{zinn2003computational}
Zinn, C.
\newblock (2003).
\newblock A computational framework for understanding mathematical discoursexy.
\newblock {\em Logic Journal of IGPL}, 11(4):457--484.

\end{thebibliography}

\end{document}